\documentclass[conference]{IEEEtran}
\IEEEoverridecommandlockouts
\usepackage{cite}
\usepackage{amsmath,amssymb,amsfonts}
\usepackage{graphicx}
\usepackage{textcomp}
\usepackage{xcolor}
\usepackage[a4paper, total={184mm,239mm}]{geometry}
\def\BibTeX{{\rm B\kern-.05em{\sc i\kern-.025em b}\kern-.08em
    T\kern-.1667em\lower.7ex\hbox{E}\kern-.125emX}}

\usepackage{algorithm}
\usepackage{algpseudocode}
\usepackage{tensor}
\usepackage{tikz}
\usepackage{pgfplots}
\pgfplotsset{compat=1.16}
\usepackage{listings}
\usepackage[inline]{enumitem}
\usepackage{xcolor}
\colorlet{punct}{red!60!black}
\definecolor{background}{HTML}{EEEEEE}
\definecolor{delim}{RGB}{20,105,176}
\colorlet{numb}{magenta!60!black}
\lstdefinelanguage{json}{
    basicstyle=\normalfont\ttfamily,
    numbers=left,
    numberstyle=\scriptsize,
    stepnumber=1,
    numbersep=8pt,
    showstringspaces=false,
    breaklines=true,
    frame=lines,
    backgroundcolor=\color{background},
    literate=
     *{0}{{{\color{numb}0}}}{1}
      {1}{{{\color{numb}1}}}{1}
      {2}{{{\color{numb}2}}}{1}
      {3}{{{\color{numb}3}}}{1}
      {4}{{{\color{numb}4}}}{1}
      {5}{{{\color{numb}5}}}{1}
      {6}{{{\color{numb}6}}}{1}
      {7}{{{\color{numb}7}}}{1}
      {8}{{{\color{numb}8}}}{1}
      {9}{{{\color{numb}9}}}{1}
      {:}{{{\color{punct}{:}}}}{1}
      {,}{{{\color{punct}{,}}}}{1}
      {\{}{{{\color{delim}{\{}}}}{1}
      {\}}{{{\color{delim}{\}}}}}{1}
      {[}{{{\color{delim}{[}}}}{1}
      {]}{{{\color{delim}{]}}}}{1},
}

\begin{document}



\title{$\theta$-Resonance: A Single-Step Reinforcement Learning Method for Design Space Exploration
\thanks{T.Q. contributed to this work during an internship at Futurewei.}
}

\author{\IEEEauthorblockN{Masood Mortazavi}
\IEEEauthorblockA{\textit{IC Lab} \\
\textit{Futurewei Technologies}\\
Santa Clara, CA, USA \\
masood.mortazavi@futurewei.com}
\and
\IEEEauthorblockN{Tiancheng Qin}
\IEEEauthorblockA{\textit{Industrial Engineering} \\
\textit{UIUC}\\
Urbana, IL, USA \\
tq6@illinois.edu}
\and
\IEEEauthorblockN{Ning Yan}
\IEEEauthorblockA{\textit{IC Lab } \\
\textit{Futurewei Technologies}\\
Santa Clara, CA, USA \\
yan.ningyan@futurewei.com}

}


\maketitle

\begin{abstract}
Given an environment (e.g., a simulator) for evaluating samples in a specified design space and a set of weighted evaluation metrics---one can use $\theta$-Resonance, a single-step Markov Decision Process (MDP), to train an intelligent agent producing progressively more optimal samples. 
In $\theta$-Resonance, a neural network $Net_{\theta}$ consumes a constant input tensor and produces a policy $\pi_{\theta}$ as a set of conditional probability density functions (PDFs) for sampling each design dimension.  
We specialize existing policy gradient algorithms in deep reinforcement learning (D-RL) in order to use evaluation feedback (in terms of cost, penalty or reward) to update $Net_{\theta}$ with robust algorithmic stability and minimal design evaluations.
We study multiple neural architectures (for $Net_{\theta}$) within the context of a simple SoC design space and propose a method of constructing synthetic space exploration problems to compare and improve design space exploration (DSE) algorithms.
Although we only present categorical design spaces, we also outline how to use $\theta$-Resonance in order to explore continuous and mixed continuous-discrete design spaces. 
\end{abstract}



\section{Introduction}
The problem of design-space exploration, whether in hardware or software, has become increasingly important as product and service cycle times have become faster and variations in specificity have increased \cite{analytic_dse, zigzag, bayesian_dse, dnn_explorer, systolic, loop, fpga_dnn, extreme_dc_asic, asic_clouds}.
In design space exploration (DSE), design evaluations often incur the greatest cost rendering brute-force sampling the worst possible exploration strategy.
A variety of optimization systems come to the rescue~\cite{open_tuner, hypermapper, sherlock, ironman_pro, ironman, delving, chip_place_rl, bayesian_dse, mvmab, erdse, single}. 
The approach, which is closest to but more elaborate than~\cite{single}, relies on two insights:\begin{enumerate*}
\item No intermediate partial design can be evaluated.
\item The problem of design space exploration can be viewed as a single-step (compound-action) Markov decision process (SSMDP).
\end{enumerate*} 

In $\theta$-Resonance, our sampling policy network learns the inter-dependencies of design dimensions online. 

The single-step, compound-action formulation enables us to
\begin{enumerate*}
\item
greatly improve the sample efficiency of our RL-guided DSE,
\item
subsume the inherent errors in artificial modeling of conditional probabilities of design moves in partial design states,
\item
avoid the compounding errors caused by inaccurate predictions, and yet
\item
make use of advanced policy gradient concepts and algorithms (e.g., PPO~\cite{schulman2017,hsu2020}) to teach our sampling agent a dynamic online policy to seek the optimal design.
\end{enumerate*}

In this report, we discuss the theoretical details of our algorithm to the extent that others can replicate the work and we provide some sample applications and explorations, leaving the more detailed discussion of its application to future reports. 



\section{Background}
\begin{figure}
\centering
\noindent\resizebox{\columnwidth}{!}{
\begin{tikzpicture}
\draw[thick,<->] (0,5) node[above]{$O_2$}--(0,0)--(5,0) node[below]{$O_1$};
\node [below left] at (0,0) {$0$};
\node [below] at (2.5,0) {${O_1}^{*}$};
\node [left] at (0,2.5) {${O_2}^{*}$};
\draw[blue](0.5,4.5)--(4.5,0.5) node[align=left, right] {$Cost^{optimal}=w_{O_1}{O_1}+w_{O_2}{O_2}$};
\draw[blue, dashed](0.5, 5.5) -- (4,2)
node[align=left,right] at (2.5, 4)
{$Cost^{suboptimal}=w_{O_1}{O_1}+w_{O_2}{O_2}$ \\ Suboptimal Constant-Cost Plane};
\draw[black](0,2.5)--(2.5,2.5)--(2.5,0);
\draw[fill, black] (2.5,2.5) circle[radius=0.1] 
node[right] at (2.6, 2.6) {Optimal Operating Point};
\draw[thick, red] (1,5) .. controls (1.65,3) and (3,1.65) .. (5,1) node[align=left, right] at (4.5,1.5) {$<O_1(X_1,...,X_D), O_2(X_1,...,X_D)>$ \\ Pareto Front};
\end{tikzpicture}
}
\caption{Simplified schematic of Pareto front, optimal point, constant-cost planes in a multi-objective scenario.}
\label{fig:pareto}
\end{figure}
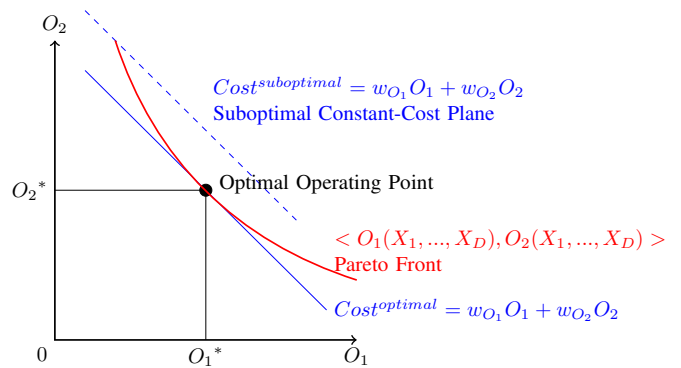
Figure~\ref{fig:pareto} schematizes (in $2$-D) the multi-objective optimization problem.
The $O_j$ is represents scarce resource used (a kind of \textit{penalty}) associated with objective $j$. 
The Pareto front of the design space represents the set of potentially achievable design points which are not ``dominated'' by any other possible design points. A design point $u$ dominates another point $v$ if all of $u$'s penalties are smaller than $v$'s. Specific weighted sum of penalties define parallel ``constant-cost planes''. 

The  major dichotomies in automated DSE are as follows:
\begin{enumerate*}
\item
Methods that model a theoretical Pareto front to guide sampling towards a better estimate of that front~\cite{hypermapper, sherlock, bayesian_dse} vs.\  methods that move sampling perpendicularly to constant-cost planes towards lower total cost~\cite{open_tuner, ironman_pro, chip_place_rl, delving, erdse}.
\item
Sampling methods which are guided by reinforcement learning~\cite{ironman_pro, chip_place_rl, delving, erdse, single} vs.\ sampling methods which are guided by Bayesian methods, genetic algorithms or heuristics~\cite{bayesian_dse, open_tuner, mvmab}. 
\end{enumerate*}

Pareto-modeling DSE approaches~\cite{hypermapper, sherlock}, model how $K$ performance objectives $O_1, O_2,...,O_K$ depend on $D$ design decisions $x_1, x_2,...,x_D$, focusing on one~\cite{hypermapper} or more~\cite{sherlock} models for the full Pareto front.
In each iteration, they sample theoretically dominating points using their model(s) and evaluate these design points and then update their Pareto-predictive models.
These methods are effectively trying to find points that lie on a $(K-1)$-dimensional \textit{surface} dominating all other design points in a $K$-dimensional objective space. 
Other approaches~\cite{open_tuner, ironman_pro, chip_place_rl, delving, erdse, single} make no specific attempt to construct this full surface. 
They either use a single objective ~\cite{open_tuner, single} or, like our method, construct a single cost based on weighted sum of objectives~\cite{ironman_pro, chip_place_rl, delving, erdse}, and focus on moving perpendicular to the parallel constant-cost planes towards the optimal point. 
(See Figure~\ref{fig:pareto}. Clearly, finding a set of points can be more sample-efficient than constructing a $(K-1)$-dimensional surface. So, a design engineer who has some notion of the relative importance of objectives can use these methods.)

\section{Proposed Method} \label{proposed}

Here, we discuss $\theta$-resonance---a method to train a sampling policy for a given design space exploration viewed as an economic search problem.
Given reward feedback, we use a policy-gradient algorithm~\cite{sutton2018reinforcement, schulman2017, hsu2020} for online improvement of our sampling policy. 

\subsection{Reinforcement Learning Conception}
The methods that rely on reinforcement learning to guide sampling and discovery of lower-cost design points on successively lower constant-cost planes~\cite{ironman_pro, chip_place_rl, delving, erdse} generally (with the exception of this work and \cite{single}) conceive of the design space exploration as a series of multi-step episodes of the form $s_0 \rightarrow x_1 \rightarrow s_1 \rightarrow x_2 \rightarrow s_2 \rightarrow \cdots \rightarrow x_D \rightarrow s_D$. Each episode begins with an incomplete design template $s_0$ and after $D$ design decisions, arrives at a terminal, ``design-complete'' state: $s_D = s_T$. 

Considering that no reward signal is available until we have evaluated the complete design, where all $D$ decisions have already been made, the only reason earlier investigations have insisted on maintaining the multi-step conception seems to be that available design moves (actions) have dependencies on partial design states $s_i$ where $0<i<D$---i.e., they depend on previous design moves, and these inter-dependencies need to be accounted for.
However, multi-step methods need to encode partial design states and use non-analytic masking techniques (e.g., in \cite{chip_place_rl, delving}) to enforce conditionality on predicted design action probabilities (policies). In the single-step MDP conception, we avail ourselves of reward orientation (in the spirit of~\cite{sutton2018reinforcement}) and avail ourselves of the capacity in our $\theta$-resonant policy network $Net_{\theta}$ to learn design-action, conditional inter-dependencies (Figure~\ref{fig:resonance}).

In short, we have come to the conclusion that we can use a single-step (compound-action) MDP ($SSMDP$) reinforcement-learning model for DSE: $s_o \rightarrow \boldmath{X} \rightarrow s_D$ where $\boldmath{X}$ is a compound action $<x_1, x_2, \cdots , x_D>$ and where the cost (penalty or reward) signal has all that is required to learn about compound-action inter-dependencies. 

Conceiving DSE as a $SSMDP$ is also a natural fit to any and all other fixed-horizon reinforcement learning problems where cost signal is only available at the end of the \textit{fixed-length} horizon---in the complete, terminal state.

\subsection{Learning Joint Distributions and Inter-Dependencies}
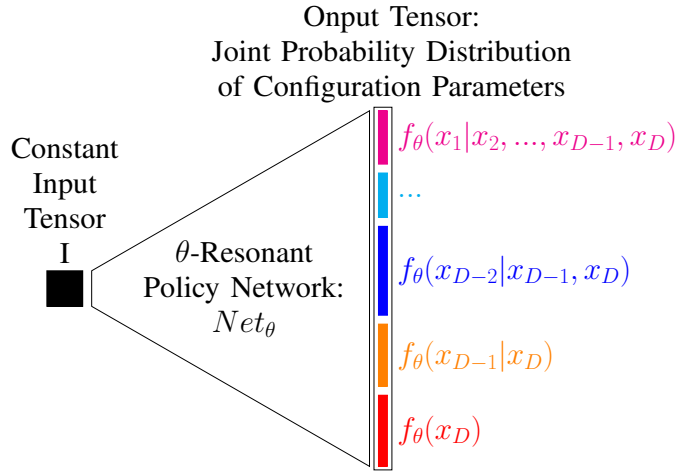
\begin{figure}
\centering
\noindent\resizebox{\columnwidth}{!}{
\begin{tikzpicture}

\tikzstyle{every node}=[font=\Huge]

\draw[thick, fill] (1,4.5) rectangle (2, 5.5)
node[align=center, above] at (1.5, 5.6)
{Constant \\ Input \\ Tensor \\ I};

\draw[thick] (2.25,4.5) -- (2.25, 5.5) -- (10, 10) -- (10, 0) -- cycle
node[align=center, font=\Huge] at (6.5,5)
{
$\theta$-Resonant \\ 
Policy Network: \\
$Net_{\theta}$
};

\draw (10.125, -0.1) rectangle (10.625, 10.1)
node[align=center, above] 
{Onput Tensor: \\ 
Joint Probability Distribution \\
of Configuration Parameters}
;

\draw[thick, red, fill] (10.25, 0) rectangle (10.5, 2) node[align=left, right]
at (10.65, 1)
{$f_{\theta}(x_D)$};	
\draw[thick, orange, fill] (10.25, 2.25) rectangle (10.5, 4)
node[align=left, right] 
at (10.65, 3.1)
{$f_{\theta}(x_{D-1}|x_D)$}; 	
\draw[thick, blue, fill] (10.25, 4.25) rectangle (10.5, 6.75)
node[align=left, right]
at (10.65, 5.5)
{$f_{\theta}(x_{D-2}|x_{D-1}, x_D)$}; 	
\draw[thick, cyan, fill] (10.25, 7) rectangle (10.5, 8.25) 
node[align=left, right]
at (10.65, 7.625)
{$. . .$};	
\draw[thick, magenta, fill] (10.25, 8.5) rectangle (10.5, 10) 
node[align=left, right]
at (10.65, 9.25)
{$f_{\theta}(x_1|x_2,...,x_{D-1}, x_D)$};

\end{tikzpicture}
}
\caption{$\theta$-Resonant Policy Network produces the joint probability density of design actions as a set of conditional probabilities in each design dimension.}
\label{fig:resonance}
\end{figure}
In $\theta$-Resonance, as shown in Figure~\ref{fig:resonance}, the sampling policy $\pi$ is generated by a trainable neural network, with parameters $\theta$: $\pi_{\theta} \: = \:  \pi(x_1...x_D | \theta , I)$. 
The constant input $I$ (e.g., a tensor of $1$'s or hot-$1$'s) represents the initial  blank slate of the design template, $s_0$, which never changes in any \textit{particular} DSE problem.
The terminal state, $s_T = s_D$, is always accomplished in a \textit{single} step that involves making all the choices $\boldmath{X}$ at once in accord with the current policy $\pi(x_1...x_D | I, \theta)$, 
where $X_1=x_1 \land X_2=x_2 \land ... \land X_D = x_D$ with random variable $X_i$ represents the $i$-th design action.

To make training and policy network size manageable, the policy $\pi(x_1....x_D | I)$ is generated as a set of conditional probabilities capturing and learning the inter-dependencies through online improvement cycles: 
Reward feedback from batches of samples is used to
update network weights $\theta$ through policy-gradient reinforcement-learning algorithms. 
Algorithm~\ref{alg:wr} gives an overview of this policy training procedure and its modules. 
The (sampling) policy ``improvement'' cycle begins with Line~\ref{alg:wr:cycle}. 
In each cycle, we seek to improve the (sampling) policy-generating neural network. 
In a sense, the policy network decodes the constant input $I$ into the joint ``action'' distribution 
$\pi(x_1...x_D | \theta , I)$, expressed in its conditional form:
\begin{equation}
\pi(x_1...x_D ; \theta , I) = \prod_{i=1}^{D} \pi_{\theta}^i = \prod_{i=1}^{D} f(x_i|x_{i+1}...x_D; \theta, I) 
\label{eq:joint}
\end{equation} 
For example, in the case of the Listing~\ref{lst:cpu_design_space}, our network will decode $I$ into a vector of length 
$2+12+3+2+3+3+4+7+13+10+5+10+5+6+7+5+7+2=106$, 
with the first $2$ elements representing the conditional discrete probabilities of the $2$ available CPU types, the next $12$ elements the conditional discrete probabilities of the $12$ available memory types, etc.

We conditionally sample each choice dimension through the neurally-predicted conditional PDFs. Since the $\theta$-Resonant policy network produces $\pi_{\theta}$ as a collection of these conditional probabilities for each dimension $i$ ($\pi_{\theta}^i$) by ``resonating'' on a constant tensor $I$, it is critical to design the policy network with good capacity for internal resonance particularly among the dimensions so that inter-dependencies can be learned through online exploration. It is by this internal resonance that the network eventually learns to produce a joint distribution for the optimal design or come to approximate it, with the distribution for optimal design tending towards $\delta(x_1^*,x_2^*,...,x_D^*)$, a multi-dimensional delta function concentrated on some optimal point $(x_1^*,x_2^*,...,x_D^*)$. To get good resonance, we have experimented with MLP and Transformer policy network types of various shapes and structure. 
\subsection{The (Sampling) Policy Optimization Algorithm}
\begin{algorithm}
\caption{$\theta$-Resonance}
\label{alg:wr}
\begin{algorithmic}[1]
\Require $<Input, Net_\theta, R_{\tau}, t_{\tau}, Sim, Obj, W>$
\\\textit{Definitions:} \hrulefill
\State $\vec{F} \equiv <f(x_1|x_2,...,x_D),f(x_2|x_3,...,x_D),...,f(x_D)>$
\State $\vec{S} \equiv <\xi_1, \xi_2,...,\xi_B>$ \Comment{Batch of Sampled Designs}
\State $\vec{O}(Sim(\xi_i)) \equiv <O_1, O_2,...,O_M>_i$ \Comment Objective Metrics  
\State $\vec{R} \equiv <R_1, R_2,..., R_B>$ \Comment{Batch of Rewards}
\State $\hat{\hat{R}} \equiv R_{running}$ \Comment Running Reward  
\\\textit{Initialization:} \hrulefill
\State $Input \gets 1$ \Comment{Constant input for the resonating network}
\State $Net(\theta_{0})$  \Comment{Initialize resonating network parameters $\theta$}
\State $\theta_{old} \gets \theta_{0}$ \Comment{$L_{e, c}(\theta_{0})$ will initiate change} \label{alg:wr:old_init}
\State $t \gets 0$
\\\textit{Policy Improvement Cycle:} \hrulefill
\While{$(R_{best sample} \leq R_{\tau}) \lor (t \leq t_{\tau})$} \label{alg:wr:cycle}
\State $\vec{F} \gets Net(\theta_{t})(Input)$
\State $\vec{S} \gets SampleFrom(\vec{F})$ \Comment{Sample}
\State $\vec{R} \gets  \vec{O}(Sim(\vec{S})) \times \vec{W}$  \Comment{Simulate}
\\\textit{Policy Gradient Updates:} \hrulefill
\State $R_{statistics} \gets RewardStatistics(\vec{R})$
\State $\hat{\hat{R}} \gets UpdateRunningReward(R_{statistics}, \hat{\hat{R}})$
\State Compute $L_{total}(\theta_{t})$ \Comment{Statistical Risk, Equation~\ref{eq:risk}}
\State $\theta_{old} \gets \theta_{t}$  \label{alg:wr:old_update}
\State $Net(\theta_{t+1}) \gets Optimize(L_{total}(\theta_{t}), Net(\theta_{t}))$ 
\State $t \gets t+1$
\EndWhile
\end{algorithmic}
\end{algorithm}
We train our policy network using policy-gradient family of D-RL algorithms with particular focus on proximal policy optimization technique~\cite{sutton2018reinforcement,schulman2017,hsu2020}. 

\subsubsection{Sampling, Simulation and Reward Assignment}
The policy network produces the joint PDF $\pi_{\theta}$ as a set of conditional PDFs, $\vec{F}$, with the relationship expressed already in Equation~\ref{eq:joint}. We use the conditional PDFs $F_i$ for each dimension $i$ to sample a batch of possible designs $\vec{S}$.
Next, each design sample $\xi_b$ in batch $B$ is simulated in parallel simulation engines.
Each result set produced by each sample simulation, $Sim(\xi_b)$, are filtered to extract objective metrics achieved by the corresponding design sample $\xi_b$ in the batch.
These objective metrics are combined, using economic weights (scarcity ``prices''), in order to arrive at a single weighted-sum, real-number reward, $R_b$ for each sample $\xi_b$: 
\begin{equation}
R_b = \sum_{j=1}^{K} w_j \cdot O_{jb}
\label{eq:weighted}
\end{equation}
Here, $O_{jb}$ is the $j$-th objective metric evaluated for the $b$-th sample in the batch. 
\subsubsection{Advantage: Reducing Variance in the Reward Signal}
Based on the reward set for the batch $\vec{R}$, we evaluate batch-level statistics in order to update a running reward $\hat{\hat{R}}$ and then compute the surrogate advantage $A_b$ for each member of the batch $\xi_b$:
\begin{equation}
\hat{\hat{R}}_{t+1} = \alpha_{renew} \cdot mean_B(\vec{R}) + (1 - \alpha_{renew}) \cdot \hat{\hat{R}}_{t}
\label{eq:running_reward}
\end{equation}
\begin{equation}
A_b = R_b - \hat{\hat{R}}
\label{eq:advantage}
\end{equation}
The surrogate advantage $A_b$ scales the adjustments to the probability of sampled design $\xi_b$ that led to reward $R_b$. 
\subsubsection{Policy Gradient}
Using policy gradient statistical risk, an optimization algorithm, e.g., SGD/Adam\cite{kingma}, and an auto-differentiation tool supporting back-propagation, we update policy network $Net_{\theta}$ which produces $\pi_{\theta}$.   
Policy gradient algorithms use the \textit{reward} signal to gradually reshape $\pi_{\theta}$ towards some optimal (sampling) policy $\pi^* \: = \: \pi_{\theta^{*}} \: = \: \delta(x_1^*,x_2^*,...,x_D^*)$.  
\subsubsection{Statistical Losses} 
The statistical risk is composed of three parts.
%
%
The conditional update loss, $L_{u,c}(\theta)$ can be computed as:
\begin{equation}
L_{u, c}(\theta) =   -  \mathop{\mathbb{E}}_{B} [ \frac{\prod_{i=1}^{D} f(x_i|x_{i+1}...x_D; \theta, I)}{\prod_{i=1}^{D} f(x_i|x_{i+1}...x_D; \theta_{old}, I)} \cdot A(\theta_{old}) ]
\label{eq:u_loss_c}
\end{equation}
Following justifications in~\cite{hsu2020}, we use $KL,rev$-divergence to regulate how rapidly the policy is allowed to change.
%
%
We use the sum of all conditional probability's $KL,rev$-divergence loss as a surrogate $KL$-regularizer:
\begin{multline}
L_{KL,rev,c}(\theta) = \\
 \beta_{KL} \sum_{i=1}^{D} D_{KL} ( f(x_i|x_{i+1}...x_D; \theta, I) ||   f(x_i|x_{i+1}...x_D; \theta_{old}, I) )
\label{eq:KL_rev_loss_c}
\end{multline}
%
%
We control exploration with a surrogate conditional entropy loss.
\begin{equation}
L_{e, c}(\theta) = - \beta_{e} . \sum_{i=1}^{D} \alpha_{i} \mathbb{H} ( f(x_i|x_{i+1}...x_D; \theta, I) )
\label{eq:e_loss_c}
\end{equation}
where $\alpha_i = log(1/d_{max})/log(1/d_i)$ normalizes relative contributions of (multi-categorical) conditional entropies\footnote{In the initial experiments reported in this paper, we set $\alpha_i = 1$.}, and $d_i$ representing the cardinality of the $i$-th (categorical) design choice dimension.
%
%
%
The statistical risk that our (sampling) policy minimizes can then be assembled as follows:
\begin{multline}
L_{total} ( \theta ) = L_{u, c}(\theta) + L_{e, c}(\theta) + L_{KL,rev,c}(\theta)
\label{eq:risk}
\end{multline}
Our algorithm includes some additional hyper-parameters discussed below. These hyper-parameters have well-defined semantics. With fast synthetic dimensionality benchmarks in the design (or index) space of a given DSE problem, the user of our system can experiment with hyper-parameter settings prior to more time-consuming and expensive simulation-bound DSE.
\subsubsection{Shaping Exploration}\label{sec:entropy}
Near the convergence horizon, too great an entropy may distract the policy into less optimal loci. 
%
We use $\beta_{e, 0}$, $\beta_{min}$ and $r_{decay}$ as hyper-parameters used to adjust entropy penalty and control the degree of exploration, starting from an initial value and decaying to a minimum final value.
\begin{equation}
\beta_{e, t} = max(\beta_{min}, \beta_{e, t-1} \cdot r_{decay})
\label{eq:shape_e_beta}
\end{equation} 
\subsubsection{Handling Simulation Anomalies}
The simulator may fail to simulate some sampled designs $\xi_b$. In fact, some design combinations will generally be physically impossible. $\theta$-Resonance accounts for these ``anomalous'' cases by assigning an appropriate reward, $R_a$, in order to discourage their generation.
\begin{multline}
R_{a} = min( \mathop{\mathbb{E}}_{B} [ R_i ] - \alpha_{a}  \lvert \mathop{\mathbb{E}}_{B} [ R_i ] \rvert  \; , \;
 \hat{\hat{R}} -  \alpha_{a}  \lvert \mathop{\mathbb{E}}_{B} [ R_i ] \rvert ) 
\label{eq:shape_anomaly}
\end{multline}
We found the structure of Equation~\ref{eq:shape_anomaly} provides for a more graceful dynamic reward assignment in anomalous cases where a simulator rejects a design.
Other more specific methods for ``anomalous'' reward assignment may be appropriate in a given scenario.
%
%
\subsubsection{Other Design Domain Types}
While we only provide categorical examples in this discussion and evaluation of our algorithm, the output of $Net_{\theta}$ in Figure~\ref{fig:resonance} can represent parameters of continuous distributions instead of categorical conditional probabilities. For example, if we want to determine macro placement as in~\cite{chip_place_rl, delving}, we can generate location as an $x-y$ coordinate through sampling two Beta distribution per macro, with each Beta distribution requiring two parameters to be generated by $Net_{\theta}$. We leave further discussion of this technique to future reports. Currently, our system automatically generates $Net_{\theta}$ based on domain requirements in the design space specification the design engineer provides. 

%
%


\section{Evaluation}
We  have tested our algorithm in various exploration scenarios in very large synthetic as well as real spaces. Artificial DSE settings are designed for rapid dimensional benchmarking and hyper-parameter configuration. We have used our system to explore designs in physical placement as well as SoC design. 
Here, for demonstration purposes, we evaluate our system against genetic algorithms (GA, e.g., combinations of normal greedy mutation, differential evolution, and uniform greedy mutation 
as implemented in~\cite{open_tuner}) in the context of two design space topologies. 
Genetic algorithms commonly start with random seeds on every run. In general practice, we fix the random seed for repeatability but to make a fair comparison, in all experiments reported here, we ran GA implementations 8 times, and we also ran our system with 8 randomly selected seeds when comparing results.

\subsection{Gem5-Aladdin Simulation}
Consider design exploration over Listing~\ref{lst:cpu_design_space} which allows for a SoC design with different x86 CPU configurations integrated with an accelerator interface. 
Given the number of options available across the $18$ choice dimensions in Listing~\ref{lst:cpu_design_space}, the number of possible designs are: $2 \times 12 \times 3 \times 2 \times 3 \times 3 \times  4 \times 7 \times 13 \times 10 \times 5 \times 10 \times 5 \times 6 \times 7 \times 5 \times 7 \times 2 = 3.4673184 \times 10^{12}$.  
In this very simple case, a limited set of benchmarks related to memory copy operations can be simulated within about $10$ seconds: A brute-force optimality search will take roughly $1.1 \times {10^6}$ years!  
\begin{lstlisting}[language=json,numbers=none, caption={A simple SoC's design space.}, label={lst:cpu_design_space},basicstyle=\footnotesize]
cpu_type:[DerivO3CPU,TimingSimpleCPU],
mem_type:[DDR3_1600_8x8,DDR3_2133_8x8,LPDDR3_1600_1x32,GDDR5_4000_2x32,HBM_1000_4H_1x128,HBM_1000_4H_1x64,WideIO_200_1x128,DDR4_2400_8x8,DDR4_2400_4x16,DDR4_2400_16x4,SimpleMemory,LPDDR2_S4_1066_1x32],
cacti_cache size(bytes):[16384, 32768, 65536],
cacti_tlb_size(bytes):[256,512],
cacti_lq_size(bytes):[128, 256, 512],
cacti_sq_size(bytes):[64, 128, 256],
sys_clock:[1GHz,2GHz,4GHz,8GHz],
num_cpus:[1,2,4,8,16,32,64],
mem-size:[1MB,2MB,4MB,8MB,16MB,32MB,64MB,128MB,256MB,512MB,1024MB,2048MB,4096MB],
l1i_size:[1kB,2kB,4kB,8kB,16kB,32kB,64kB,128kB,256kB,512kB],
l1i_assoc:[1,2,4,8,16],
l1d_size:[1kB,2kB,4kB,8kB,16kB,32kB,64kB,128kB,256kB,512kB],
l1d_assoc:[1,2,4,8,16],
l2_size:[128kB,256kB,512kB,1024kB,3048kB,4096kB],
l2_assoc:[1,2,4,8,16,32,64],
l3_size:[512kB,1024kB,2048kB,4096kB,8192kB],
l3_assoc:[1,2,4,8,16,32,68],
cacheline_size:[32,64]
\end{lstlisting}


%
%
We evaluate sampled designs $xi_b$ using the Gem-5-Aladdin SoC simulator\cite{gem5a}.  
In this DSE task, we compare the number of samples the automatic explorer generates for evaluation purposes before it reaches a certain level of cost (penalty, negative reward). 
\subsubsection{Architectural and Batching Variations}
\begin{figure}
\centering
\noindent\resizebox{\columnwidth}{!}
{

\begin{tikzpicture}
  \begin{loglogaxis} 
[
	legend style={font=\tiny,  align=right,  legend pos=south west}, 
    title={Sample Efficiency,  Gem5-Aladdin SoC DSE Test}, 
    xlabel={Unique Samples Visited}, 
    ylabel={$\Delta\:=\:R_{target}\:-\:R_{bs}$}, 
    label style={font=\small}, 
    tick label style={font=\footnotesize},  
] 


\addplot+[const plot, mark=diamond, thick, color=black] coordinates {
(1, 326058.6 - 116100 )
(5, 317960.8 - 116100 )
(9, 280886.8 - 116100 )
(38, 246986.9 - 116100 )
(45, 244809.9 - 116100 )
(71, 237970.9 - 116100 )
(115, 191086.9 - 116100 )
(152, 173127.9 - 116100 )
(156, 162760.9 - 116100 )
(208, 162412.2 - 116100 )
(217, 150350.9 - 116100 )
(225, 131473.9 - 116100 )
(273, 128450.9 - 116100 )
(274, 116486.9 - 116100 )
(337, 116195.9 - 116100 )
}
;    
\addlegendentry{T12.67M-8-8-4}


\addplot+[const plot, mark=o, mark options={solid}, dotted, thick, color=blue] coordinates {
(1, 721059.0 - 116100 )
(5, 368226.6 - 116100 )
(8, 299088.7 - 116100 )
(10, 288397.5 - 116100 )
(27, 195952.9 - 116100 )
(284, 184082.9 - 116100 )
(454, 180937.9 - 116100 )
(713, 177831.0 - 116100 )
(787, 163849.9 - 116100 )
(828, 160539.9 - 116100 )
(1146, 156626.9 - 116100 )
(1241, 142644.9 - 116100 )
(1292, 140055.9 - 116100 )
(1410, 117591.9 - 116100 )
(1767, 116602.9 - 116100 )
(2239, 116486.9 - 116100 )
(2272, 116327.9 - 116100 )
(2545, 116195.9 - 116100 )
}
;    
\addlegendentry{MLP13.72M-8-8-4}

\addplot+[const plot, mark=*, mark options={solid}, dotted, thick, color=blue] coordinates {
(1, 721059.0 - 116100 )
(5, 368226.5 - 116100 )
(8, 299088.7 - 116100 )
(10, 288397.5 - 116100 )
(28, 195952.9 - 116100 )
(280, 185022.9 - 116100 )
(451, 180937.9 - 116100 )
(491, 171189.9 - 116100 )
(785, 163849.9 - 116100 )
(1203, 163158.9 - 116100 )
(1233, 160615.9 - 116100 )
(1375, 158160.9 - 116100 )
(1396, 152967.9 - 116100 )
(1404, 136183.9 - 116100 )
(1761, 132212.9 - 116100 )
(2095, 124140.9 - 116100 )
(2480, 123725.9 - 116100 )
(2538, 123400.9 - 116100 )
(2829, 122458.9 - 116100 )
(2923, 116602.9 - 116100 )
(3245, 116327.9 - 116100 )
(3515, 116195.9 - 116100 )
}
;   
\addlegendentry{MLP8.88M-8-8-4}

\addplot+[const plot, mark=x, mark options={solid}, dotted, thick, color=blue] coordinates {
(1, 721059.0 - 116100 )
(5, 368226.5 - 116100 )
(8, 299088.7 - 116100 )
(10, 288397.5 - 116100 )
(27, 195952.9 - 116100 )
(265, 185022.9 - 116100 )
(483, 157535.9 - 116100 )
(1456, 149893.9 - 116100 )
(2160, 147908.9 - 116100 )
(2304, 147773.9 - 116100 )
(2513, 130660.9 - 116100 )
(5172, 128450.9 - 116100 )
(5340, 128430.9 - 116100 )
(7614, 123400.9 - 116100 )
}
;   
\addlegendentry{MLP3.58M-8-8-4}

\end{loglogaxis}

\end{tikzpicture}

}
\caption{Sample Efficiency,   Gem5-Aladdin SoC DSE Test: Policy Network Variations. $R_{bs}$ is the best seen reward up to the number of samples. The $R_{target}\:=\:-116100$ is set for display purposes. It highlights how close varying runs get to 116195.9, which is the best ever seen reward.}
\label{fig:arch_var_gem5a}
\end{figure}
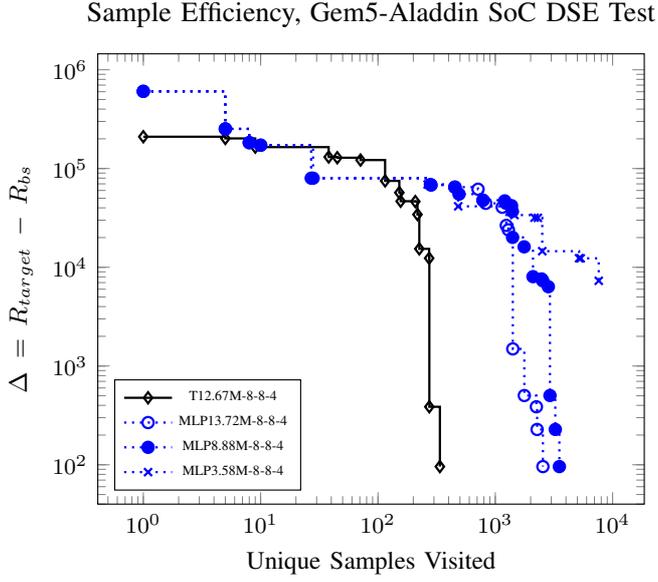
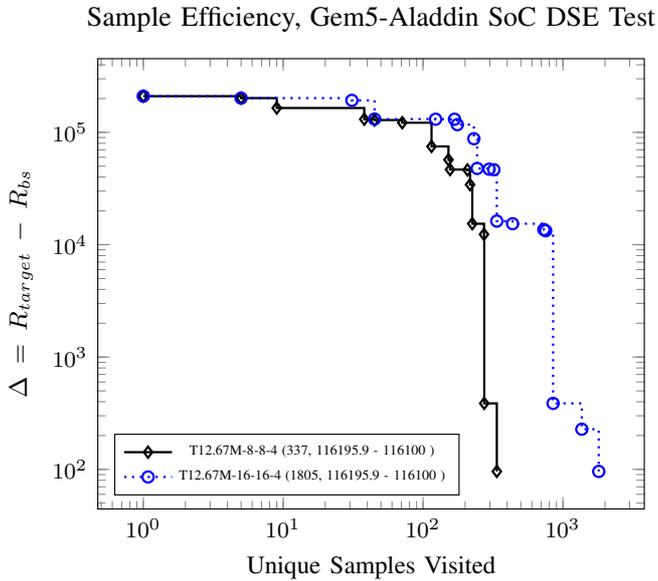
\begin{figure}
\centering
\noindent\resizebox{\columnwidth}{!}
{

\begin{tikzpicture}
  \begin{loglogaxis} 
[
	legend style={font=\tiny,  align=right,  legend pos=south west}, 
    title={Sample Efficiency,  Gem5-Aladdin SoC DSE Test}, 
    xlabel={Unique Samples Visited}, 
    ylabel={$\Delta\:=\:R_{target}\:-\:R_{bs}$}, 
    label style={font=\small}, 
    tick label style={font=\footnotesize},  
] 


\addplot+[const plot, mark=diamond, thick, color=black] coordinates {
(1, 326058.6 - 116100 )
(5, 317960.8 - 116100 )
(9, 280886.8 - 116100 )
(38, 246986.9 - 116100 )
(45, 244809.9 - 116100 )
(71, 237970.9 - 116100 )
(115, 191086.9 - 116100 )
(152, 173127.9 - 116100 )
(156, 162760.9 - 116100 )
(208, 162412.2 - 116100 )
(217, 150350.9 - 116100 )
(225, 131473.9 - 116100 )
(273, 128450.9 - 116100 )
(274, 116486.9 - 116100 )
(337, 116195.9 - 116100 )
}
;    
\addlegendentry{T12.67M-8-8-4 (337, 116195.9 - 116100 )}

\addplot+[const plot, mark=o, mark options={solid}, dotted, thick, color=blue] coordinates {
(1, 326058.6 - 116100 )
(5, 317960.8 - 116100 )
(31, 308910.8 - 116100 )
(45, 247487.9 - 116100 )
(123, 247075.9 - 116100 )
(168, 247040.5 - 116100 )
(176, 233273.9 - 116100 )
(231, 203930.9 - 116100 )
(244, 163845.9 - 116100 )
(297, 163142.9 - 116100 )
(321, 162587.0 - 116100 )
(337, 132327.9 - 116100 )
(437, 131473.9 - 116100 )
(731, 129752.9 - 116100 )
(753, 129414.9 - 116100 )
(851, 116486.9 - 116100 )
(1366, 116327.9 - 116100 )
(1805, 116195.9 - 116100 )
}
;    
\addlegendentry{T12.67M-16-16-4 (1805, 116195.9 - 116100 )}

\end{loglogaxis}

\end{tikzpicture}

}
\caption{Sample Efficiency,   Gem5-Aladdin SoC DSE Test: Batching Variations. $R_{bs}$ is the best seen reward up to the number of samples. The $R_{target}\:=\:-116100$ is set for display purposes. It highlights  how close varying runs get to 116195.9, which is the best ever seen reward.}
\label{fig:batching_var_gem5a}
\end{figure}
Figure~\ref{fig:arch_var_gem5a} shows the impact of architectural variations. (Note: $T12.67M-8-8-4$ means Transformer with $12.67M$ run with PPO batch/mini-batch/epoch configuration 8/8/4, and $MLP$ stands for multi-layer perceptron.) 
These results are found using our default random seed.
With the exception of the smallest MLP, all both MLP and Transformer architectures lead to the design with the least penalty found by any of our systems (minus 116100, for better display). 
The Transformer~\cite{allyouneed}  policy network seems most efficient in learning joint dependencies and converges to the best design with the least number of samples.
Figure~\ref{fig:batching_var_gem5a} provides a view of some PPO batching variations when using the Transformer policy network. Larger sampled batched under the same policy cause slower learning rates but are also generally more stable.
\subsubsection{Comparison with Genetic Algorithms}
\input{gem5a_sample_efficiency_log_log.tex}
We used a Transformer~\cite{allyouneed} implementation of our policy network when comparing to GA implementation.
Figure~\ref{fig:se_gem5a} compares $\theta$-Resonance's DSE sample efficiency with GA.
$\theta$-Resonance always converges to the best configuration visited by either itself or GA implementation with lower deviations. 
Only 7 of the 8 runs with GA implementation converge, with a wider variation, and one of the 8 runs remains at some significant distant from empirical optimal when we stopped the run, implying far larger variation in convergence.
We see this trend also in our synthetic tests. 
 
\subsection{Using Synthetic Benchmarks}
\input{tt_solution_sample_efficiency.tex}
One of the main challenges in comparing DSE systems is the unavailability of common benchmarks.
To study large design spaces we have devised a class of very fast exploration-space sample-efficiency benchmarks that can be used for two purposes:
\begin{enumerate*}
\item to compare system performance of alternate algorithms in particular design space topologies, and 
\item to explore best-practice values for hyper-parameters when using those algorithms in a given design space topology. 
\end{enumerate*} 
As a very simple example, let us consider a synthetic decision space of $20$ dimensions with $64$ \textit{categorical} choices available in each dimension. (This simple synthetic example is for demonstration purposes only.)
This space contains $1.329228 \times 10^{36}$ possible design configurations.
%
%
Unbeknownst to the automatic explorer an arbitrarily point is selected as a ``black-box'' point in the exploration space. This selection allows an artificial distant reward to be used as a quick black-box feedback to the automatic explorer.
Figure~\ref{fig:se_set_20d64_solution} compares GA implementation with $\theta$-Resonance in this synthetic DSE task.
Again, $\theta$-Resonance can get arbitrarily close to the optimal point (and indeed always finds the optimal point).
The combination GA algorithms (as implemented in \cite{open_tuner}) never find the optimal point although they are dapable of getting close to it.
This offers another confirmation for the advantage of the single-step MDP ($\theta$-Resonance) approach.
Similar experiments in larger artificial spaces have corroborated these results. 


\subsection{Algorithmic Variations}
We also tried several other algorithmic variations and ablation studies within our own algorithm all of which cannot be reviewed here given the scope of this report.
Here, we simply note that we did try the clip loss which is popular in most uses of PPO~\cite{schulman2017}. 
However, $KL_{rev, c}$-divergence (see Equation~\ref{eq:KL_rev_loss_c}) behaved better and more robustly for us as has also been substantiated and discussed in great detail by Hsu\cite{hsu2020}.
It is worth noting that we conducted all of our algorithmic-internal variations and ablation studies using the synthetic benchmarks---and recommend this approach for rapid prototyping and evaluation of automatic DSE algorithmic variation. 
%


%

\section{Discussion}

In this section, we present a brief discussion of some of the intuitions  and insights based on our algorithm and on its evaluation given in the earlier sections above. 

\subsection{Single-Step Reinforcement Learning for Design Space Exploration}

In the language of RL, we are dealing with the special case of a single (compound-choice) step. This problem can also be conceived of as a multi-variate, multi-armed bandit.  However, conceiving it as a single-step RL opens the door to the use of policy-gradient RL algorithms. 

While single-step arrival at a terminal state is rarely used in developing RL theory or applications, the theoretical developments and the algorithms still apply, and we take advantage of this by carefully simplifying the existing policy-gradient algorithms, and consequently our implementation of them, in order to allow for what we have come to call $\theta$-Resonance, a single-step, compound-action RL approach, which is also identical with a single-step markov decision process. 

There are multi-step conceptions \cite{chip_place_rl} but they imply a need for conditional action probabilities in a partial design as they incrementally complete the design. 

Since designs cannot be evaluated until they are ``complete'' in some sense, the only advantage of multi-step method is that it allows use of existing RL tools but it also runs into the problem of non-analytic conditional probabilities which these multi-step methods try to solve through artificial discretization of action domains and subsequent masking of action network's probability predictions for those parts of the action space which have, in some sense, become ``impossible'' in each given step. In a sense, they try to deploy and make use of scanty prior knowledge. (See discretization discussion and Figure 1 in \cite{chip_place_rl}, and the critique of handling essentially continuous spaces as discrete spaces in \cite{hsu2020}.) 

Our experience with our algorithm tells us such intra-design choice dependencies (effectively, conditional probabilities) can also be learned in a single step conception as long as penalties and rewards are properly defined. As Sutton has emphasized
\cite{sutton2018reinforcement}, the use of reinforcement learning requires a careful definition of reward. Hence, for example, our proposal for engineering a reasonable dynamic reward to handle designs which simulators find impossible, as given in Equation~\ref{eq:shape_anomaly}.

In this way, we off-load and delimit the designer's prior knowledge entirely from the algorithm to the reward signals generated by the simulator. We free our algorithm from handling such prior knowledge which other approaches try to artificially incorporate algorithmicly by imposing a multi-step, stateful conception built on partial design states which are hard to interpret and which force the algorithm designer to use artificial methods (e.g., grids and masks, as in \cite{chip_place_rl, delving}) to compute conditional action probabilities.

In particular, and in contrast to the claim in \cite{delving} that reward-orientation cannot help with learning macro overlap (given, as another argument for artificially construction of conditional probabilities in the output of multi-step action network that depends on masks and incomplete design state tracking), we were able to learn to avoid macro overlap using our single-step method in the problem of macro placement in the physical design of a chip\footnote{This work has not been included here and will be presented in subsequent papers.}.
We were able to accomplish this through (negative) reward assignment to the undesirable overlap of macros in conjunction with other rewards that need to be weighted. As noted above, the only rationale in \cite{chip_place_rl} for multi-step treatment seems to be the avoidance of overlap---and this leads to the use of grids and masks. The problem that \cite{delving} had in learning overlap probably have to do with its limitations in collecting high-fidelity data given that it is dealing with long (but artificial and constant length) episodes---where we are actually dealing with a single step, compound action to get to the terminal state of a complete design in a definite number of choices. Our experiments in this regard are beyond the scope of the present paper but they have proven to us that single-step RL in DSE is far more sample-efficient than multi-step RL in DSE. 

Close to publication, the only other work we found that uses single-step formulation for design space exploration is~\cite{single} where single-step method is used to explore some low-dimensional design spaces. Our formulation is far more general. We clearly establish an algorithm for attacking large-dimensional problems, describe an approach to designing and generating policy networks, define the statistical losses used as well as un-biasing running reward and rewards for anomalous cases and use fast (and synthetic) dimensional benchmarks to identify regions of effective hyper-parameter settings.  

In short, DSE is naturally a single-step compound-action RL, i.e., episodes have definite terminal states within a fixed and definite horizon for any given design exploration problem involving a design template with a fixed set of design dimensions left to be determined through automatic DSE.
From an optimization point of view, DSE is essentially a single-step process because interpreting partial designs proves to be unnecessary. 
Any partial design becomes complete through a fixed number of choices---and only a complete design can be evaluated. 

\subsection{Resonance and Expected PDF Shape in the Optimality Limit}
\label{app:delta}
In the optimality limit, if entropy bonus for exploration is eventually removed or if the explorer is penalized for over-exploring in the limit approaching optimality, the joint action probability (i.e., the sampling ``policy'' $\pi_{\theta}$), in continuous decision dimensions, will tend towards a product of $\delta$ functions at the optimal point(s) as discovered by the ``resonating'' policy network\footnote{The analogy in discrete domains is that the probability density will concentrate, in each decision dimension, on a single optimal value or on some definite subset of such values. In the case of multiple optimal values, the joint probability converges to the normalized sum of a product of delta density functions, each product representing one optimal design point.}. 
\begin{multline}
\pi(x_1...x_D ; \theta^{*} , I) = \prod_{i=1}^{D} \delta(x_i^*) 
\label{eq:deltas}
\end{multline}
At such an optimality limit, which we also observed in those experiments where we gradually removed the entropy bonus, the network should generally produce no new samples but only the optimal one it has discovered---with the policy network arriving at its complete and final resonant state.

Inclusion of a minimum entropy (as in Equation~\ref{eq:shape_e_beta}) implies continuing search in at best some noisy region near the optimality limit. 

\subsection{Implicit Learning of Conditional Dependencies}
Of course, as displayed in Figure~\ref{fig:resonance}, we model the joint action probability $\pi_{\theta}$ as the product of a set of conditional probabilities represented by the output of our policy network\footnote{For the relationship between conditional and joint probabilities, see~\cite{feller,parzen}.}. 

Our inspiration for this type of modeling was that the internal nodes in the network can implicitly learn the inter-dependencies of various choice dimensions and are able to generate a useful representation of joint probabilities of design decisions  as their conditional probability components over a fixed and definite set of design dimensions for any given design problem.

After having constructed our system and near publication, we noticed that Bengio and Bengio~\cite{bengio_bengio} had also used this concept of joint-conditional learning in the intermediate tiers.
However, their system can only be used for learning the joint-conditional distribution of an input data-set whereas our $\theta$-Resonance algorithm learns a sampler for optimum designs.

In~\cite{bengio_bengio}, the input to a neural network varies over an input dataset and the ``predicted'' output is the distributional parameters of the dataset. 

Given the constant input in $\theta$-Resonance representing the ``blank'' state of the partial design template, our policy network's output varies based on changes we make to its weights. 
Our policy network can only produce a different output through policy gradient updates of its weights in response to external reward feedback signal coming in the form of a non-differentiable (with respect to the weights $\theta$) reward which is produced by evaluating the set of design decisions suggested by that neural network's output (as the joint probability of the optimum, compound decision). The design evaluation must be conducted through the use of a simulator or through real-world tests---i.e., in some RL ``environment'' whether simulated or real-world.
In other words, while \cite{bengio_bengio} is about distributional summarization of a dataset, our $\theta$-Resonance is about exploring a design space to find an optimal design. 
Nevertheless, it is interesting that others had also used the intuition the internal layers of a neural network can learn conditional dependencies when outputs are joint distribution parameters.

\subsection{Categorical, Continuous, Ordinal and Mixed Domains}
The design space of the example in Listing~\ref{lst:cpu_design_space} is composed of a number of possible discrete (categorical) choices.
However, out method will also be applicable to cases where choice dimensions include continuous dimensions of choice or mixed discrete-continuous dimensions of choice where some dimensions involve continuous and others discrete choices. Here, regarding continuous design decision dimensions, we merely note that one can use analytic continuous probabilities that are most suitable for a given problem at hand---e.g., through the Beta distribution for bounded continuous dimensions or the Gaussian distribution for unbounded continuous dimensions\footnote{Other analytic probabilities are, of course, possible.}. 
In such cases, the conditional action probabilities for continuous dimensions would be generated by our policy networks as the corresponding analytic probability density function's parameters. 

Multi-step, stateful methods that attempt to capture prior knowledge on conditional action probabilities lead to unnecessary discretization and ad-hoc masking of output probabilities. The reader may consult with \cite{hsu2020} for some further discussion of the flaws of discretization in continuous domains and use of continuous parametric distributions. It should be clear intuitively that discretization leads to loss of reward-based policy gradient fidelity\footnote{The proof of this claim is beyond the scope of this paper but we have observed this affect empirically.}. 

The case of ordinal decision dimensions can also be included but they often lead to the complementary use of other methods of attack, which we do not address here. 
Our focus has been to examine pure sample-efficiency of the exploration system, which becomes more critical when no models (not even interpolation models) are available.

\section{Conclusion}

Our exploration strategy, $\theta$\textit{-Resonance}, is distinct from earlier approaches in the sense that
\begin{enumerate*}
\item
it conceives of design space exploration as a single-step, compound-action Markov decision process, and
\item
it prescribes a trainable and direct sampling policy which implicitly learns the inter-dependencies of the choices in each design dimension.
\end{enumerate*}
%
%

In this first report on $\theta$-Resonance, besides describing our RL-based policy-gradient algorithm, we demonstrate its use in real-world and in synthetic dimensional test environments 
and show that our transformer policy network is more effective in learning join inter-dependencies among design dimensions in the real-world scenario.
%

We have shown how a policy-gradient algorithm can be modified to suit our one-step, compound-action scheme and have shown its competitive sample efficiency in design space exploration.
%
%

Finally, we have discussed a number of intuitions and ideas related to our proposal which can be used for applying and extending it in a variety of design exploration problems.
%
   
%
%
%
%
%

\section*{Acknowledgment}

We would like to thank our management at Futurewei Technologies, and in particular Dr. Liang Peng for his encouragement, support and guidance throughout the conception and development of this research project. Ziru Chen, an earlier intern on this project, worked with us on continuous choice domains (related to the problem of macro-placement) the intuitions from which we have briefly discussed in this work.

\bibliographystyle{IEEEtran}
\bibliography{theta_resonance}

\appendix
The software implementation for this algorithm along with the system for synthetic benchmarks, search domain specification and simulator integration layers 
and samples will be subsequently released, potentially on github, for the purpose of further research and integration. It may include other design evaluation simulators and optimization problems.
It is relatively easy to implement new policy networks, new algorithms and integrate simulators for a vareity of DSE problems within our $\theta$-Resonance framework.

\end{document}